\documentclass[journal]{IEEEtran}

\ifCLASSINFOpdf
\else
\fi

\usepackage{times}
\usepackage{epsfig}
\usepackage{graphicx}
\usepackage{amsmath}
\usepackage{amssymb}
\usepackage{multirow}
\usepackage{overpic}
\usepackage{tikz}
\usepackage{pgfplots}
\pgfplotsset{compat=1.17} 
\usepackage{dsfont}
\usepackage[switch]{lineno}

\usepackage{caption}
\usepackage{subcaption}
\usepackage{amssymb}
\usepackage{pifont}
\usepackage{pict2e}

\usepgfplotslibrary{external}
\newcommand{\cmark}{\ding{51}}%
\newcommand{\xmark}{\ding{55}}%
\newcommand{\best}[1]{\textbf{#1}}
\newcommand{\correct}[1]{\textbf{#1}}

\def\eg{\emph{e.g.}}
\def\ie{\emph{i.e.}}
\def\etal{\emph{et. al}}
\DeclareMathOperator*{\argmax}{argmax}
\usepackage[pagebackref=false,breaklinks=true,letterpaper=true,colorlinks,bookmarks=false]{hyperref}

\begin{document}
\title{Temporal Action Segmentation with High-level Complex Activity Labels}

\author{Guodong Ding, Angela Yao
\thanks{Guodong Ding and Angela Yao are with the School of Computing, National University of Singapore, Singapore, 117418. E-mails: dinggd@comp.nus.edu.sg, ayao@comp.nus.edu.sg}%
\thanks{Manuscripts received August 15, 2021.}}

\markboth{Journal of \LaTeX\ Class Files,~Vol.~14, No.~8, August~2015}%
{Shell \MakeLowercase{\textit{et al.}}: Bare Demo of IEEEtran.cls for IEEE Journals}
\maketitle

\begin{abstract}
The temporal action segmentation task segments videos temporally and predicts action labels for all frames. Fully supervising such a segmentation model requires dense frame-wise action annotations, which are expensive and tedious to collect.

This work is the first to propose a Constituent Action Discovery (CAD) framework that only requires the video-wise high-level complex activity label as supervision for temporal action segmentation. The proposed approach automatically discovers constituent video actions using an activity classification task. Specifically, we define a finite number of latent action prototypes to construct video-level dual representations with which these prototypes are learned collectively through the activity classification training. This setting endows our approach with the capability to discover potentially shared actions across multiple complex activities. 

Due to the lack of action-level supervision, we adopt the Hungarian matching algorithm to relate latent action prototypes to ground truth semantic classes for evaluation. We show that with the high-level supervision, the Hungarian matching can be extended from the existing video and activity levels to the global level. 
The global-level matching allows for action sharing across activities, which has never been considered in the literature before. Extensive experiments demonstrate that our discovered actions can help perform temporal action segmentation and activity recognition tasks.
\end{abstract}

\begin{IEEEkeywords}
Temporal Action Segmentation, Weakly Supervised Learning, Hungarian Matching, Prototype Learning, Activity Recognition
\end{IEEEkeywords}

\IEEEpeerreviewmaketitle

\section{Introduction}

\newcommand{\reducedstrut}{\vrule width 0pt height .9\ht\strutbox depth .9\dp\strutbox\relax}
\newcommand{\col}[2]{%
  \begingroup
  \setlength{\fboxsep}{0pt}%
  \colorbox[HTML]{#1}{\scriptsize \reducedstrut  #2\/}%
  \endgroup
}

\newcommand{\colo}[2]{%
  \begingroup
  \setlength{\fboxsep}{0pt}%
  \colorbox[HTML]{#1}{\reducedstrut  #2\/}%
  \endgroup
}
\begin{figure}[t]
    \centering
    \begin{overpic}[width=0.5\textwidth]{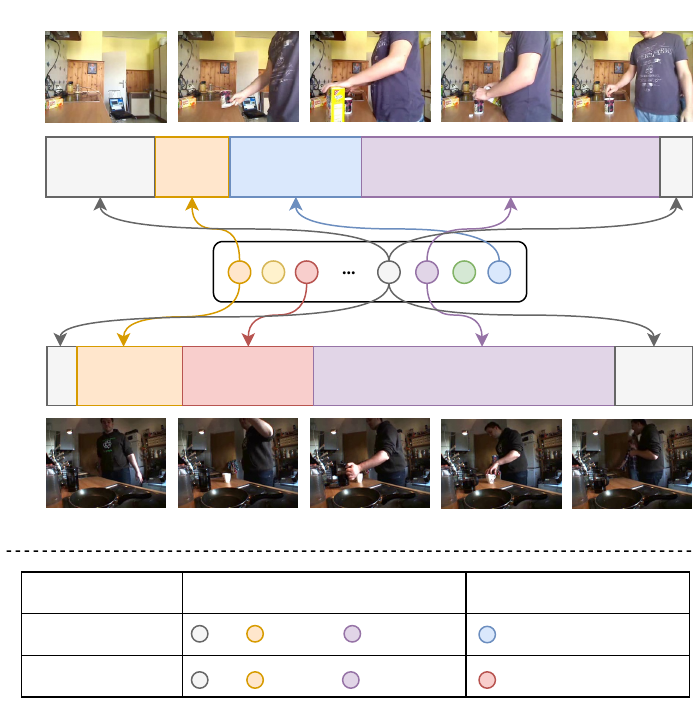}
    \put(43,97){\scriptsize (chocolate) milk}
    \put(12,75){\scriptsize SIL}
    \put(22.5,75){\scriptsize take\_cup}
    \put(34.4,75){\scriptsize spoon\_powder}
    \put(67,75){\scriptsize pour\_milk}
    \put(94.3,75){\scriptsize SIL}
    
    \put(6.8,45){\scriptsize SIL}
    \put(14,45){\scriptsize take\_cup}
    \put(29,45){\scriptsize pour\_coffee}
    \put(61,45){\scriptsize pour\_milk}
    \put(91,45){\scriptsize SIL}
    \put(50,24){\scriptsize coffee}

    \put(0,42.5){\rotatebox{90}{\scriptsize Action}}
    \put(2.5,40.8){\rotatebox{90}{\scriptsize Segments}}
    \put(0,72.5){\rotatebox{90}{\scriptsize Action}}
    \put(2.5,70.8){\rotatebox{90}{\scriptsize Segments}}
    \put(0,30.5){\rotatebox{90}{\scriptsize Video}}
    \put(2.5,29.8){\rotatebox{90}{\scriptsize Frames}}
    \put(0,85.5){\rotatebox{90}{\scriptsize Video}}
    \put(2.5,84.8){\rotatebox{90}{\scriptsize Frames}}
    \put(78, 60){\scriptsize Action Set}

    \put(4, 14.5){\scriptsize Complex Activity}
    \put(27, 14.5){\scriptsize Shared Composing Actions}
    \put(68, 14.5){\scriptsize Activity Exclusive Actions}
    \put(4, 9){\scriptsize (chocolate) milk}
    \put(30, 9){\col{F5F5F5}{SIL}}
    \put(38, 9){\col{FFE6CC}{take\_cup}}
    \put(52, 9){\col{E1D5E7}{pour\_milk}}
    \put(71, 9){\col{DAE8FC}{spoon\_powder}}
    
    \put(30, 2.5){\col{F5F5F5}{SIL}}
    \put(38, 2.5){\col{FFE6CC}{take\_cup}}
    \put(52, 2.5){\col{E1D5E7}{pour\_milk}}
    \put(71, 2.5){\col{F8CECC}{pour\_coffee}}
    \put(4, 2.5){\scriptsize coffee}
    
    \end{overpic}
    \caption{Temporal action segmentation of two video instances from Breakfast. It shows that each video instance can be seen as a composition of actions; also, certain actions like `{\colo{FFE6CC}{take\_cup}}' and `\colo{E1D5E7}{pour\_milk}' appear in both complex activity ``(chocolate) milk'' and ``coffee'' when performing action segmentation, while some actions are activity exclusive, \eg, `\colo{DAE8FC}{spoon\_powder}' and `\colo{F8CECC}{pour\_coffee}', as listed in the table.} 
    \label{fig:intro}
\end{figure}
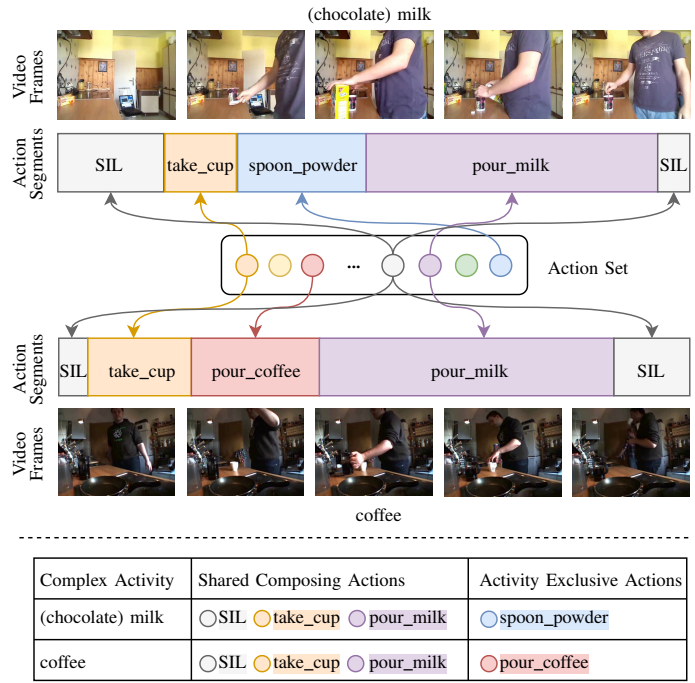

In the past few years, much of the effort in video understanding has focused on action recognition on trimmed videos~\cite{roy2018unsupervised,yu2019weakly,zhao2020universal,li2018unified,moniruzzaman2021human}. 
Standard action recognition targets the classification of short, pretrimmed clips of single actions. 
In contrast, procedural tasks and instructional videos are highly challenging to work with since they tend to be minutes long and contain multiple actions that are related to each other through sequence dynamics. 
The temporal action segmentation task, aimed at temporally segmenting videos and predicting frame-wise action labels, has attracted the increasing attention of the research community, and a variety of supervised learning methods have been developed~\cite{farha2019ms,li2020ms,chen2020action,ishikawa2021alleviating,wang2020boundary}. 
However, annotating every frame in videos is highly labour-intensive. As such, other lines of work learn with weak forms of supervision~\cite{laptev2008learning,richard2018action,richard2017weakly,richard2018neuralnetwork} or entirely without supervision~\cite{sener2018unsupervised,kukleva2019unsupervised,vidalmata2021joint}.

We are interested in discovering and segmenting the full set of constituent steps, \eg, `take\_cup', `pour\_milk', `pour\_coffee', `spoon\_powder', in procedural videos given only the type of task as a label, \eg, ``(chocolate) milk'' and ``coffee''. 
We refer to these steps as \emph{`actions'} and the procedural task as a \emph{``complex activity''} and illustrate an example in Fig.~\ref{fig:intro}.
Previous attempts~\cite{sener2018unsupervised,kukleva2019unsupervised,vidalmata2021joint,elhamifar2019unsupervised} have been dedicated to addressing such an action discovery and segmentation problem in an unsupervised fashion. 
Notably, these methods handle a collection of videos of the same complex activity and thus use the high-level complex activity labels individually. 

Discovering actions within the isolated complex activity is non-ideal.
First and foremost, it cannot find communal or shared actions across multiple complex activities, \eg, `take\_cup' and `pour\_milk' can be part of making both ``(chocolate) milk'' and ``coffee'' in Fig.~\ref{fig:intro}. 
Secondly, it scales with the number of complex activities, even though the number of constituent or composing actions is fixed. 
For a complete understanding, we posit that action discovery should be maattractedlobal basis across multiple complex activities. 

\begin{table*}[!htb]
\centering
\caption{Comparison of supervisory signals and evaluation prerequisites in temporal action segmentation. Full supervision provides frame-wise dense action labels. The timestamp setting provides an ordered list of actions per video with corresponding exemplar frames. Evaluation of such two settings relies on network predictions without any pre-steps. Action lists or set supervision do not provide exemplars. Evaluation under this setting is based on the best-matched action list searched with maximum sequence posterior for each test video. Our setting uses the same amount of supervision information as the unsupervised counterpart. However, ours uses all complex activity videos simultaneously compared to unsupervised ones, which use them one at a time. Since no action-level signals are provided, it is necessary to perform Hungarian matching before evaluation.}
\label{tab:sup}
\begin{tabular}{l|c|c|c|c|c}
\hline
\multirow{2}{*}{} & \multirow{2}{*}{Full~\cite{farha2019ms}} & \multicolumn{3}{c|}{Weak} & \multirow{2}{*}{Unsupervised~\cite{sener2018unsupervised,kukleva2019unsupervised,vidalmata2021joint}} \\ \cline{3-5}
 &  & \multicolumn{1}{c|}{Timestamp~\cite{li2021temporal}} & \multicolumn{1}{c|}{Action list/set~\cite{richard2018action,fayyaz2020sct}} & Ours &  \\ \hline
Action labels & \begin{tabular}[c]{@{}c@{}}Dense \\ frame-wise\end{tabular} & \multicolumn{1}{c|}{\begin{tabular}[c]{@{}c@{}}ordered list + \\ exemplar frames\end{tabular}} & \multicolumn{1}{c|}{\begin{tabular}[c]{@{}c@{}}(ordered) list, \\ union set\end{tabular}} & - & - \\ \hline
Activity labels & - & \multicolumn{1}{c|}{-} & \multicolumn{1}{c|}{-} & \begin{tabular}[c]{@{}c@{}}All activities \\ en masse\end{tabular} & \begin{tabular}[c]{@{}c@{}}One activity \\ at a time\end{tabular} \\ \hline
\begin{tabular}[c]{@{}l@{}}Evaluation \\ prerequisite\end{tabular} & - & - & \multicolumn{1}{c|}{\begin{tabular}[c]{@{}c@{}}maximum sequence \\ posterior\end{tabular}} & \begin{tabular}[c]{@{}c@{}}Hungarian \\ matching\end{tabular} & \begin{tabular}[c]{@{}c@{}}Hungarian\\  matching\end{tabular} \\ \hline
\end{tabular}
\end{table*}

\begin{figure}[]
    \centering
    \begin{overpic}{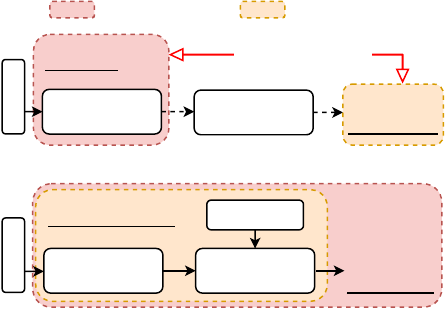}
        \put(27,1){\small (b) Embedding w/ Prototypes}
        \put(28,37){\small (a) Embedding + Clustering}
        
        \put(22, 72){\scriptsize : supervised task}
        \put(65, 72){\scriptsize : unsupervised task}
        
        \put(1.7,49){\rotatebox{90}{\scriptsize Inputs}}
       
        \put(10,61){\scriptsize pretext task}
        \put(15,51){\scriptsize Embedding}
        \put(17,47.5){\scriptsize Network}
        
        \put(55,62){\scriptsize \textbf{\textcolor{red}{feature-to-task gap}}}
        
        \put(50,49){\scriptsize Clustering}
        
        \put(84,51){\scriptsize action}
        \put(80,47.5){\scriptsize segmentation}
        
        \put(1.7,14.5){\rotatebox{90}{\scriptsize Inputs}}
        
        \put(12,26){\scriptsize action segmentation}
        \put(15,15){\scriptsize Embedding}
        \put(17,11.5){\scriptsize Network}
        
        \put(52,15){\scriptsize Affinity}
        \put(52.5,11.5){\scriptsize Matrix}
        
         \put(51,26){\scriptsize Prototypes}
         
        \put(82,15){\scriptsize activity}
        \put(80,11.5){\scriptsize recognition}
        
    \end{overpic}
    \caption{Comparison of feature-task discrepancies between clustering approaches and our proposed framework. (a) Pretext task guided feature embedding learning creates a gap between the segmentation tasks. (b) No feature-to-task gap in our framework as the feature learning is based on the action segmentation results embodied in the activity recognition task. }
    \label{fig:gap}
\end{figure}
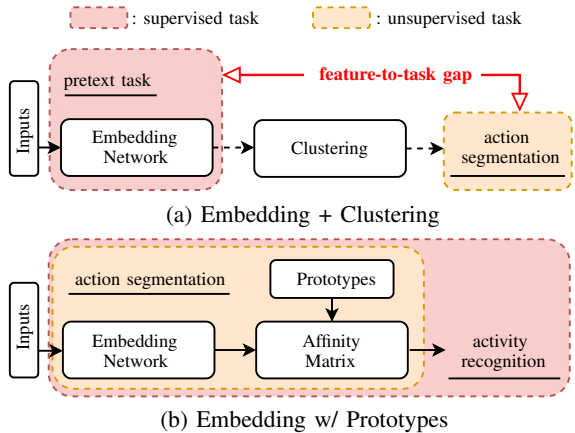

On a separate note, without labels of any kind, unsupervised learning methods typically resort to clustering-based solutions. 
Existing methods use various forlabor-intensive~\cite{sener2018unsupervised,kukleva2019unsupervised,vidalmata2021joint} or optimization~\cite{elhamifar2019unsupervised,bojanowski2014weakly}. 
In particular,~\cite{kukleva2019unsupervised,vidalmata2021joint} aim to learn better visual representations and temporal attributes via embedding networks before clustering. 
However, learning embedding with pretext tasks, \eg, timestamp prediction in~\cite{kukleva2019unsupervised}, creates a `feature-to-task' gap between the embedding and the action segmentation itself, as demonstrated in Fig.~\ref{fig:gap}(a), and little has been done to address this gap in this research community.

Hence, we are motivated to both discover a global set of actions and reduce the above `feature-to-task' gap in the temporal action segmentation task.  
Inspired by works~\cite{wu2019long,feichtenhofer2019slowfast} that use frame-wise action labels to boost the performance of activity recognition, we adopt a reverse strategy and employ complex activity labels to help discover constituent actions. 
We start by representing video instances with a set of learnable action prototypes. 
Then, we train the video classification network with activity labels and learn the prototypes simultaneously. 
Lastly, we perform frame recognition or action segmentation based on the frame-wise similarity with respect to the prototypes. 
In other words, we build on top of the action segmentation results to represent a video sequence and perform activity classification (see Fig.~\ref{fig:gap}(b)) for training. Such a setting avoids the `feature-to-task' gap and is highly intuitive since the complex activity can be regarded as a set of actions with some weak temporal order to serve a purpose or achieve a specific goal.

Our problem setup occupies a unique position on the supervision spectrum. Table~\ref{tab:sup} provides a detailed comparison between different supervision signals. First of all, we consider our work a weakly supervised method since we train with activity labels.  
Unlike other weakly supervised works with more vigorous forms of action-level supervision, \eg, (ordered) lists~\cite{kuehne2017weakly,richard2017weakly} or union sets~\cite{richard2018action,fayyaz2020sct} of actions, our work only uses video-level activity annotations.
Compared to other unsupervised works~\cite{sener2018unsupervised,kukleva2019unsupervised,vidalmata2021joint} that handle a collection of same-activity videos at a time, our approach uses the entire corpus of videos simultaneously. Our framework implicitly utilizes the same amount of (label) information as the previous works. However, it discovers composing actions on a broader scope, which we consider `global', allowing for action sharing across multiple complex activities. 

It is noteworthy that without any action labels in either our weakly supervised or existing unsupervised settings, the prototypes (clusters) do not have semantic labels to evaluate performance properly. 
The common practice is to perform Hungarian matching between prototypes and ground truth classes and find the best matching label. Depending on the scope of the two matching bodies, there are also hierarchies of Hungarian matching protocols to apply, ranging from \textbf{per-video} (\cite{elhamifar2019unsupervised,sarfraz2021temporally}) and \textbf{per-activity} (\cite{sener2018unsupervised,kukleva2019unsupervised,vidalmata2021joint}) to the \textbf{global} scope of the entire video set. To the best of our knowledge, our work is the first to perform Hungarian matching and evaluate temporal action segmentation on the global level.

The main contributions of our work are fourfold: 
\begin{itemize}   
    \item To the best of our knowledge, we are the first to perform action segmentation that allows for shared actions in a global setting, and we also show that our approach can be effectively adapted for per-activity evaluation; 
    \item We propose an action discovery framework that discovers constituent actions by performing activity classification on dual video representations derived from prototypes, bridging the `feature-to-task' gap in previous works;
    \item We analyze existing Hungarian matching protocols and further generalize the protocol to the `global' level to complete the matching hierarchy. The clear division helps to establish the standard for performance comparisons in the research field of temporal action segmentation; 
    \item Our proposed method achieves competitive unsupervised action segmentation performance and demonstrates unprecedented activity recognition performance on the Breakfast Actions dataset.
\end{itemize}

The rest of the paper is organized as follows. 
We first review the related works for our temporal action segmentation task in Section~\ref{sec:related}. 
Then, in Section~\ref{sec:hun}, we provide a clear division and discussion regarding the different levels of Hungarian matching protocols applied in existing unsupervised works and generalize the protocol to the global level. 
Section~\ref{sec:approach} provides the details of our proposed Constituent Action Discovery (CAD) framework.
The experimental settings and the results and analysis are explained in Section~\ref{sec:exp} and Section~\ref{sec:results}, respectively. Finally, we conclude our paper in Section~\ref{sec:conclusion}.

\section{Related Work} 
\label{sec:related}

Temporal action segmentation is a fast-growing area where early work typically requires videos that are fully annotated with action classes and their start and end points~\cite{richard2016temporal,lea2017temporal}.
Recently, there has been a growing interest in decreasing the amount of supervision using accompanying narrations~\cite{Sener_2015_ICCV,fried2020learning} and ordered or unordered lists of actions~\cite{richard2018neuralnetwork,kuehne2017weakly,fayyaz2020sct}. 
More preferable are algorithms requiring no supervision~\cite{sener2018unsupervised,kukleva2019unsupervised,vidalmata2021joint}. 
Currently, the majority of these works assume that frame- or snippet-level features are extracted and made available for the segmentation task, such as improved dense trajectories (IDT)~\cite{wang2013action} and I3D~\cite{carreira2017quo}. 

\textbf{Fully Supervised} approaches have been demonstrated to provide high-quality and accurate temporal segmentation with sufficient frame-wise labels. 
Previous works~\cite{farha2019ms,lea2017temporal,huang2020improving} have focused on architecture developments to capture and model the long-range temporal dependencies. 
Lea \etal~\cite{lea2017temporal} were the first to use temporal convolution networks (TCN) for segmentation. 
MS-TCN~\cite{farha2019ms} builds a cascade network by stacking multiple stages to refine the segmentation using dilated temporal convolutions progressively. 
A recent emerging idea is improving the existing segmentation algorithms by modelling the temporal relations between actions~\cite{huang2020improving,tirupattur2021modeling}, decreasing the differences
between the feature spaces of videos from different environments~\cite{chen2020action} and refining the segmentation at boundaries~\cite{wang2020boundary}.

\textbf{Weakly Supervised} approaches receive a list of actions or use complementary textual data as supervision. Such supervision is cheaper as action boundaries are no longer required. 
A common approach is using an ordered list of actions called transcripts.
~\cite{richard2017weakly} iteratively trains an RNN model to align video frames to the given actions.  
~\cite{huang2016connectionist} proposes a connectionist temporal classification-based approach for aligning the transcripts with video frames with consistency constraints. 
Using a weaker form of supervision, ~\cite{richard2018action} proposes a probabilistic model to find the action segments given unordered lists of actions, called action sets.
Similarly, ~\cite{fayyaz2020sct} uses action sets and learns the action correspondence and length by imposing a temporal consistency loss on the frame and snippet-based predictions.
Several works use instruction narrations for segmentation~\cite{Sener_2015_ICCV,fried2020learning}. Although narrations are straightforward to obtain, as they are freely available with videos, the success of these approaches depends heavily on the alignment between the narrative text and the visual data. Meanwhile, in our setting, we only utilize high-level complex activity labels, which are much easier to obtain.

\textbf{Unsupervised} learning-based approaches have recently received increasing  attention~\cite{sener2018unsupervised, kukleva2019unsupervised,vidalmata2021joint,aakur2019perceptual}. One line of work targets key-frame localization in videos~\cite{elhamifar2019unsupervised,bojanowski2014weakly}. Another growing line of work targets segmentation with minimal supervision by only requiring the complex activity label as supervision. Such supervision does not require action-level annotations but partitions videos based on their complex activity labels. 
Sener \etal~\cite{sener2018unsupervised} propose an iterative discriminative-generative approach that alternates between learning action representations and modeling their temporal structure using the generalized Mallows model. Meanwmodelingte{kukleva2019unsupervised} learns continuous temporal embeddings of frame-wise features, which are then clustered and used to decode videos based on ordered clusters. A very recent follow-up work on unsupervised temporal action segmentation task is ASAL~\cite{li2021action} which added a temporal order classification task of sampled action clips together with an alternating learning scheme to enhance the feature learning.
These methods all perform segmentation on a collection of videos with the same complex activity. 
The main difference between our work and these unsupervised works is that we can identify action classes across multiple complex activities, which we refer to as shared actions.
The closest to our work is the work of Kukleva~\etal~\cite{kukleva2019unsupervised}, who extended their model to a `global' setting by first partitioning videos into complex activities via bag of words clustering before performing action clustering. 
We note, however, that their action clustering is still performed on a per-pseudo-activity basis, and thus they are not capable of dealing with shared actions across complex activities like our proposed work. 

\textbf{Prototype Learning.} Finding prototypes and using them as exemplars to perform a particular task is common in areas such as image classification~\cite{yang2018robust,allen2019infinite} and image retrieval~\cite{wen2016discriminative}. In such works, the common practice is to define prototypes by averaging over sample representations per class in a latent feature space. 
Considering it is hard for a single prototype to capture the intra-class representation variances, there have also been works investigating multiple prototypes per class~\cite{allen2019infinite,movshovitz2017no,yang2018robust}. However, the sample embedding changes during training, so any prototypes represented as a mean vector of these samples are also unstable. This has led to the chicken-or-egg dilemma and caused the actual feature locations in the embedding space to be more challenging to obtain. Compared to these works, our framework chooses not to manipulate input sample representations in the feature space; instead, we formulate the prototypes as a fixed set of trainable parameters that can be optimized through learning a classification task.

\section{Hungarian Matching For Temporal Action Segmentation}
\label{sec:hun}
In unsupervised image classification with no class labels, such as deep clustering~\cite{caron2018deep}, it is necessary to first discover the one-to-one relationship between clusters and classes and then evaluate the classification performance. 
Such relationship discovery is made through the Hungarian matching algorithm, where the matching degrees (ratio of overlaps) are summed over all cluster-class pairs to solve for the optimized assignment~\cite{li2006relationships}.

Similarly, in unsupervised temporal action segmentation, the standard is to use the Hungarian algorithm for the one-to-one matching. 
Given the frame corpus $X$ from $N$ clusters and ground truth labels set $Y$ of $M$ action classes, Hungarian matching relates $N$ clusters to $M$ semantic labels by finding the best matching $\widehat{\mathcal{A}} \subset \{0,1\}^{N\times M}$, defined as:
\begin{multline}
\label{eq:hungarian}
    \widehat{\mathcal{A}} = \argmax_{\mathcal{A}} = \sum_{n,m} \mathcal{A}_{n,m} \cdot I(X_n, Y_m), \\
    \text{s.t.} \quad |\mathcal{A}| = \min(N,M)
\end{multline}
where $X_n$ denote the frames in cluster $n$, and $Y_m$ denote the frames with action label $m$. $\mathcal{A}_{n,m}$ is the indicator value for binding $n$ to $m$. $I(X_n, Y_m)$ calculates the number of frames with class label $m$ that appear in cluster $n$.

As indicated by Eq.~\eqref{eq:hungarian}, the result of Hungarian matching primarily bases on the scope of clusters $X_{1:N}$ and action classes $Y_{1:M}$. 
Depending on the different matching scopes of $N$ and $M$, we conclude three levels of Hungarian matching for temporal action segmentation without action-level supervision, \ie, video-level, activity-level and global-level. %

First, \textbf{video-level} matching~\cite{aakur2019perceptual,sarfraz2021temporally} confines the matching bodies to be the found action clusters and the ground truth actions of a single given video. This matching is repeated for all videos in a test set. This matching level evaluates a method's ability to partition a video sequence into individual actions. %
For the video level matching, the segmentation model cannot associate the action clusters of the same semantic meaning between different video sequences, even if the two video instances are of the same complex activity. As the simplest setting, video-level matching often produces the highest performance. 

Then, \textbf{activity-level} matching is to associate discovered action clusters to their semantic labels given a set of videos performing the same complex activity. The activity-level protocol is the most widely accepted in existing unsupervised works~\cite{sener2018unsupervised,kukleva2019unsupervised,vidalmata2021joint} as their approaches only work with a collection of the same activity videos and repeat the process on each remaining activity class. One detriment of processing activity classes individually is that it does not allow for action sharing across activities. %

Lastly, \textbf{global-level} matching is to compare on the entire dataset the action cluster outputs against the complete set of ground truth class labels.  
Global-level matching is the most challenging case as intra- and inter-activity action relations must be contemplated to find the optimal alignment. %
We note that in~\cite{kukleva2019unsupervised}, the authors reported `global' matching results across complex activities. However, their `global' setting is not equivalent to the above-described one. 
They can be distinguished based on whether shared actions across complex activities are allowed or not. 
We tally their setting and provide a detailed comparison in Sec.~\ref{subsec:unknown}.

\begin{table}[]
\centering
\caption{Model learning requirements comparison at different levels of Hungarian matching for action segmentation.}
\label{tab:hungarian}
\begin{tabular}{l|l|c|c|c}
\hline
& level & \multicolumn{1}{|c}{\begin{tabular}[c]{@{}c@{}}intra-video\\ discrimination\end{tabular}} & \multicolumn{1}{|c}{\begin{tabular}[c]{@{}c@{}}intra-activity \\ association\end{tabular}} & \multicolumn{1}{|c}{\begin{tabular}[c]{@{}c@{}}inter-activity \\ association\end{tabular}} \\ \hline
- & Video    &       \cmark      &  \xmark   &  \xmark   \\
Unsupervised &Activity &       \cmark      &  \cmark   &  \xmark   \\
Ours &Global   &       \cmark      &  \cmark   &  \cmark   \\

\hline
\end{tabular}

\end{table}

Consequently, levels of matching scopes bring escalated challenges to the model design and call for more robustness in the learning paradigms for the action segmentation task.  Table~\ref{tab:hungarian} summarizes the differences between the learning requirements of a desirable action segmentation model in all the above three matching protocols.
Generally speaking, the task becomes more challenging with the rising matching hierarchy. 
For the lowest video-level matching, disambiguation of actions within a single video,    
\ie, \emph{intra-video discrimination}, alone is sufficient to address the problem. 
On top of the discriminability of actions, the model should simultaneously learn \emph{intra-activity association} to enable activity-level matching.
Global matching is only applicable at the highest level when the model learns to incorporate \emph{inter-activity association} and establish action correspondences across activities. 

\begin{figure*}[h]
    \centering
    \includegraphics[width=\textwidth]{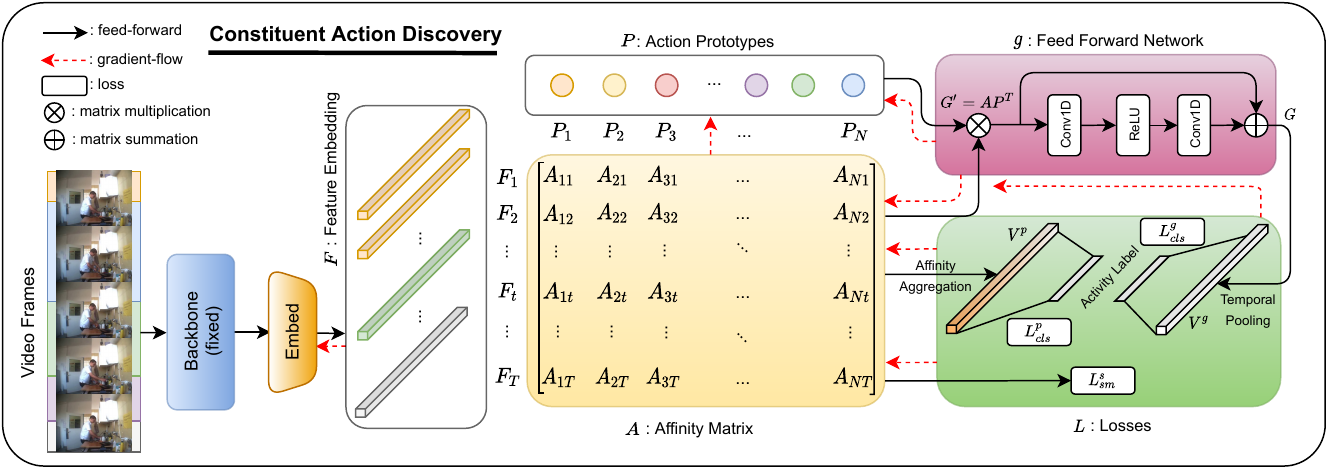}
    \caption{The overall architecture of our proposed Constituent Action Discovery (CAD) framework. The input to the framework is the pre-computed frame-wise feature from a fixed backbone network. After an embedding module, feature embedding $F$ and actions prototypes $P$ are combined to obtain affinity matrix $A$. Visual representation $V^g$ is obtained from $G$ by going through a feed-forward network $g$. Prototype representation $V^p$ is the affinity aggregation over the frames. Two video-level cross-entropy losses $L_{cls}^p$ and $L_{cls}^g$ are imposed on $V^p$ and $V^g$, respectively, for activity recognition. An extra smoothing loss $L_{sm}^s$ is imposed along the temporal dimension of affinity matrix $A$.}  
    
\label{fig:archi}
\end{figure*}

We note that any model learned at a higher level is only \textbf{downward compatible} and can be adjusted to be evaluated at a lower level but not vice versa.
Even though downward evaluation is feasible, the level at which these actions are discovered determines which aspect of the algorithm should be evaluated; therefore, the results are not directly comparable across levels. 
To the best of our knowledge, our work is the first to work at the highest global level of matching. Even though we also report results by adapting evaluation at the activity level, our discovery of actions is still global. Hence, it is still more challenging than existing unsupervised works, as discussed above and compared in Table~\ref{tab:hungarian}.

\section{Constituent Action Discovery (CAD)}
\label{sec:approach}
We present in Fig.~\ref{fig:archi} an overview of our proposed Constituent Action Discovery (CAD) framework, which is aimed at discovering a set of prototypes via training a complex activity classification network with a dual video representation design.

\subsection{Task Definition}
Given a collection of video sequences belonging to $C$ complex activities, each video annotated with a complex activity label $y \in [1, C]$, our goal is to relate each video frame, indexed by $t$, to an action label $n$ out of $N$ possible actions.  
The $N$ actions are constituent steps shared amongst the $C$ complex activities. 
The objective of our proposed approach is to learn a compilation of representations $\mathbf{P}=\{P_n\}_{n=1}^N$, which we designate as action prototypes, such that each $P_n$ well-characterizes a distinctive composing action.
Used together, the prototypes $\mathbf{P}$ should best match as many video frames as possible to the underlying set of action labels. The best match found via Hungarian matching, therefore, serves as the temporal action segmentation result.

\subsection{Dual Video Representations}
\label{subsec:dual}
For a given video, we define for each video frame feature $F_t$ at time $t$ a corresponding latent representation $G_t$.  $G_t$ is based on a mapping of the weighted summation of the $N$ prototypes, \ie:
\begin{equation}\label{eq:prototypesum}
    G_t = g(G_t'; \theta_a), \quad\text{where} \quad G_t' = \sum_{n=1}^N A_{t,n} \cdot P_n.
\end{equation}
\noindent In the above equation, the mapping $g(\cdot)$ is parametrized by $\theta_a$ while the weight $A_{t,n}$ is the affinity between $F_t$ and $P_n$. An intuitive interpretation of $G_t'$ in Eq.~\eqref{eq:prototypesum} is to consider it as the re-constructed representation of frame $F_t$ by an affinity-weighted combination of prototypes in $\mathbf{P}$.
We define 
$D_{t,n}$ as some distance between $F_t$ and $P_n$, \ie: 
\begin{equation}\label{eq:distance}
    D_{t,n} = d(F_t,P_n),
\end{equation}
where $d(\cdot)$ can be for example a Euclidean distance.  
The affinity distance for a $t$-th frame is defined as one minus a temporally normalized distance: 
\begin{equation}
    D^{'}_{t,n} = 1-\frac{D_{t,n}-[D_{t}]_{\text{min}}}{[D_{t}]_{\text{max}}-[D_{t}]_{\text{min}}},  \qquad D^{'}_{t,n} \in [0,1]
    \label{eq:distance_norm}
\end{equation}
where $D_{t} \in \mathbb{R}^N$ is a distance vector between $F_{t}$ and all the prototypes $P$ and $[\cdot]_{\text{min}}$ and $[\cdot]_{\text{max}}$ return the minimum and maximum values of the vector. 
Afterwards, we normalize $D^{'}_{n,t}$ with respect to all the prototypes so that they form an affinity distribution: 
\begin{equation}
    A_{t,n} = \frac{D^{'}_{t,n}}{\sum_{n=1}^N D^{'}_{t,n}}
\end{equation}
The more similar a frame $F_t$ is to a prototype, the closer its affinity value $A_{t,n}$ is to 1.
Thus, the affinity matrix $A \in \mathbb{R}^{T\times N}$ represents the similarity between a video sequence and all learned prototypes.  
Each row of $A$, denoted as $A_t$, represents the similarity of that frame to all prototypes.  

\textbf{Prototype Representation $V^p$.} Based on the affinity matrix, we can thus define a time-aggregated video prototype representation as $V^p$, where
\begin{equation}\label{eq:prototype_aggregation}
    V^p = \sum_{t=1}^T A_{t}.
\end{equation}
\noindent $V^p$ accumulates the evidence of prototypes over the entire video or input sequence. 
Previous weakly supervised work ~\cite{fayyaz2020sct} adopted a global max-pooling over the temporal dimension to represent the entire video sequence.  
We posit that such an operation simply ensures the existence of specific prototypes but disregards the frequency of action occurrences, which is also essential. 
Aggregation with summation over time considers both action (prototype) occurrence and frequency, thereby allowing our framework to capture the underlying action distribution. 
We empirically show this finding in Sec.~\ref{subsec:dist}.

\textbf{Visual Representation $V^g$.}  In addition to the prototype representation, we define a redundant {visual representation} $V^g$ by averaging over time the latent frame-wise representations $G_t$, \ie: 
\begin{equation}\label{eq:visual_averaging}
     V^g = \frac{1}{T}\sum_{t=1}^T G_t.
\end{equation}
Compared to prototype representation $V^p$, $V^g$ averages the feature presentations of all video frames and is designed to capture and summarize more visual cues from the frames, which are essential in recognizing complex activities. 

Based on the dual representations $V^p$ and $V^g$, we can estimate the complex activity $y$ via mappings $f_p$ and $f_g$, parameterized by $\theta_p$ and $\theta_g$ respectively: 
\begin{equation}
    \hat{y}^p = f_p(V^p; \theta_p), \qquad \hat{y}^g = f_g(V^g; \theta_g), 
\end{equation}
where $\hat{y}^p \in \mathbb{R}^C$ and $\hat{y}^g\in \mathbb{R}^C$ are the normalized activity probabilities. 
The visual and prototype representations are heterogeneous representations of a single video; both should be discriminative with respect to the complex activity on its own. 

\subsection{Loss Functions}
Our proposed approach discovers action prototypes $\mathbf{P}$ by considering a cross-entropy loss between the estimated complex activity $\hat{y}$ and given video labels:
\begin{equation}\label{eq:ce}
    L_{cls} = - \sum_{j=1}^C y_j \log(\hat{y}_j)+ (1-y_j) \log(1-\hat{y_j}),
\end{equation}
where $y$ is the one-hot vector representation of the complex activity label.  
We apply the same loss for both $\hat{y}^p$ and $\hat{y}^g$ and denote the two as $L_{cls}^p$  and $L_{cls}^g$, respectively. 
Unlike previous action segmentation works~\cite{farha2019ms,chen2020action} that impose a frame-wise cross-entropy loss, our classification loss is computed on a per-video basis.

Considering that actions occurring in an activity video should be temporally contiguous, we further adopt from~\cite{farha2019ms} a smoothing term called Truncated Mean Squared Error (T-MSE).  
Generally, this loss is applied to the per-frame actions in a supervised framework to help alleviate over-segmentation. 
While we do not have action labels, we thus impose the same loss on the affinity matrix $A$ to set temporal transition constraints on frame affinities. 
The loss is formulated as follows:
\begin{equation}\label{eq:tmse}
    L_{\text{sm}}^s = \frac{1}{TN}\sum_{t,n}\tilde{\Delta}_{t,n}^2, \;\;\;
    \tilde{\Delta}_{t,n} = \begin{cases}
    \Delta_{t,n} &: \Delta_{t,n} \le \tau\\
    \tau &: \text{otherwise}
    \end{cases},
\end{equation}
\begin{equation}
    \Delta_{t,n} = \left|\log A_{t,n} - \log A_{t-1,n}\right|.
\end{equation}

Summing the three loss terms together, we get:
\begin{equation}\label{eq:finalloss}
    L=\alpha L_{cls}^p + (1-\alpha)L_{cls}^g + \lambda L_{\text{sm}}^s,
\end{equation}
where $\alpha$ is a weighting hyperparameter for the prototype and visual representation. 
In our experiment, we simply set $\tau=4$ and $\lambda=0.15$ as per MS-TCN~\cite{farha2019ms}.

\subsection{Inference and Decoding}
With our framework, we can perform action segmentation with the discovered prototypes on either the global level or activity level with a simple adaptation. 

\textbf{Global.} For segmentation on a global basis, recall that the affinity vector $A_{t}$ is the normalized similarities between a frame at time $t$ and all prototypes $\mathbf{P}$,  a na\"{i}ve way to determine the frame labeling $l_t$ is to simply find the prototype with the highest affinity: 

\begin{equation}\label{eq:labeling}
    l_t = \underset{n}{\mathrm{argmax}} A_{t,n}.
\end{equation}

The above label $l$ relates each video frame to a discovered action prototype. To give the prototypes semantic meaning, we can use Hungarian matching to establish a one-to-one mapping to the action labels. 
Because the prototype discovery (and therefore Hungarian matching) is done across all the activities of all complex activities, we refer to this as the \emph{`global'} setting. This setting allows frames from different activity videos to have the same action label, thus enabling a shared set of prototypes.

\textbf{Activity.} Our approach is flexible and can be adapted for segmentation within single complex activities. This would put us in line with previous unsupervised methods~\cite{sener2018unsupervised,kukleva2019unsupervised,vidalmata2021joint}, which also perform discovery within a single complex activity. 
To adapt the segmentation for specific activities, we first reduce the affinity matrix to preserve only the top $N', N'<N$ most occurring labels in $l_{1:T}$ obtained from Eq.~\eqref{eq:labeling} over all videos from the same class to derive an activity-specific affinity matrix $A' \in \mathbb{R}^{T\times N'}$. $N'$ can be set either to be the same for all classes or simply to the ground truth action number per class. Afterwards, we relabel all the frames using the same equation as Eq.~\eqref{eq:labeling} but replace $A$ with $A'$. This relabeling process can be efficiently done by inserting a masking operation.

We observe, however, that the transitions between prototypes in $A'$ can be very noisy between consecutive frames. Therefore, we propose applying a simple smoothing along the temporal dimension by convolving with a Gaussian kernel: 
\begin{equation}
    \hat{A}' = A' * G(t,\sigma), 
\end{equation}
where $\sigma$ is the standard deviation of Gaussian kernel $G$. We find that $\sigma\!=\!5$ works well (see Table~\ref{tab:ablation}). Such a smoothing approach has also proved effective in~\cite{ishikawa2021alleviating,du2022fast}.

We follow the method proposed in~\cite{kukleva2019unsupervised} to perform Viterbi decoding on such sequences; we generate a sequence ordering $\mathbf{O}=\{O_j\}_{j=1}^{N'}, O_j \in [1,N]$ for all the $N'$ selected actions by calculating and sorting the average timestamps for each. The frame likelihood needed for decoding is expressed via Bayes' Rule:
\begin{equation}
    p(x_t|n) \propto \frac{p(n|x_t)}{p(n)} = \frac{\hat{A}'_{t,n}}{p(n)},
\end{equation}
where the posterior $p(n|x_t)$ is directly represented by the affinity value $\hat{A}'_{t,n}$.  For simplicity, we define the class prior $p(n)$ as a uniform distribution over all actions. A similar assumption has also been made in~\cite{li2021action}. During decoding, a frame indexed at $t$ can either keep the same label as that of the frame at $t-1$ (for example, $O_j$) or take on the label of the next label ($O_{j+1}$) observed in the predefined ordering $\mathbf{O}$.

\section{Experimental Setting}
\label{sec:exp}
\subsection{Datasets}
We evaluate our approach using two datasets: Breakfast Actions~\cite{kuehne2014language} and YouTube Instructional Videos~\cite{alayrac2016unsupervised}. Both datasets have videos with complex activity labels and action segment labels that can be used to train our constituent action discovery framework. Note that we do not use any action labels during training; they are only used for testing. 

\textbf{Breakfast Actions} is a large-scale dataset of 52 people performing ten different complex cooking activities. The number of composing actions for the entire dataset is 48; for each activity, it varies from 5 to 14. Shared actions between activities are quite common, and 13 of the 48 actions are used in at least two complex activities. The length of each video is highly dependent on the type of task and ranges from 30 seconds to a few minutes. The actions are contiguous with each other without any intermediate background frames, although the beginning and ends of the sequences are marked as background. 

\textbf{YouTube Instructional Videos} has five instructional activities, with 30 videos each of ``making coffee'', ``changing a car tire'', ``CPR'', ``jumping a car'', and ``potting a plant''. Different from Breakfast, some of the videos are produced in that they may be edited and or pieced together from several shots with different viewpoints. The videos in this dataset have longer temporal spans, and a significant portion of the frames is background.  
There are a total of 47 actions, but unlike Breakfast, the actions in these activities are not shared. Despite finding global actions as one of our primary motivations, it is not a requirement for our framework, so we treat the experiments on this dataset as a special case.

\textbf{Features.} To ensure a fair comparison with other works, we report experimental results on Fisher vector (FV) representation of improved dense trajectories~\cite{wang2013action} features and I3D~\cite{carreira2017quo} features for Breakfast. 
For YouTube Instructions, we use the same feature from~\cite{alayrac2016unsupervised}, which is the concatenation of the bag of words appearance feature from VGG16 network and the motion feature by the histogram of optical flows.

\subsection{Evaluation Metrics}
For evaluation, \textbf{Mean over Frames (MoF)} is reported on both Breakfast and YouTube to indicate the percentage of frames in the sequence that are correctly labeled over all the frames of videos assigned. \textbf{F1 score} is the average of the harmonic mean of precision and recall over sampled segments and is reported for YouTube Instructions to compare with previous works~\cite{sener2018unsupervised,kukleva2019unsupervised,vidalmata2021joint}. Note that these metrics are calculated based on the Hungarian matching results to report the best possible scores, as no action-level annotations are used during learning.

\subsection{Implementation Details}
We implement our model using Pytorch~\cite{paszke2017automatic}. 
To obtain our initial feature $F_t$, we apply a 1D convolution with a kernel size of 1 to reduce the dimensionality of the input features. $F_t$ is reduced to 1024 and 20 for I3D and Fisher vector, respectively, from their original 2048 and 64 on Breakfast. While on YouTube, $F_t$ is reduced to 512 from its original 3000. Due to the high ratio of backgrounds on YouTube, we follow ~\cite{kukleva2019unsupervised} and define a background ratio hyper-parameter $\eta\!=\!0.75$, where only $1-\eta$ percent of frames that are closest to one particular prototype are kept, and the rest are treated as background. Correspondingly, we also exclude the background when reporting the results for YouTube.

We implement the latent mapping $g(\cdot)$ (see Eq.~\eqref{eq:prototypesum}) as a simple feed-forward network of two convolutions with a residual connection to stabilize and speed up the learning process. We set our initial learning rate as 0.001 and optimize with Adam. 
We train on Breakfast with a total of 240 epochs and use 120 on YouTube Instruction Videos, both with a batch size of 8 and $N\!=\!50$ prototypes. 

\begin{table}[t]
\centering
\caption{Ablation study on Breakfast. FV denotes Fisher vector. The remaining results are reported based on I3D. $K$ denotes the number of prototypes being used. `max' denotes the maximum number of ground truth actions per activity.}
\label{tab:ablation}
\begin{tabular}{l|c|c|c|c}
\hline
Hungarian  &   K      & Decoding & Gaussian & MoF (\%)   \\ \hline
global (FV) & 50           & \xmark    & \xmark    & 10.9 \\
global &    50           & \xmark      & \xmark     & 19.2 \\\hline
activity  &    5          & \xmark         & \xmark         & 34.0 \\ 
activity  & max    &\xmark        & \xmark        & 28.9 \\ \hline
activity  & max  & \cmark       & \xmark        & 38.7\\ 
activity  & max  & \cmark      & $\sigma$=3      & 48.8 \\ 
activity  &max   & \cmark      &$\sigma$=5      & \best{53.1} \\ 
activity  & max   & \cmark     & $\sigma$=10     & 52.0 \\ \hline
\end{tabular}

\end{table}

\section{Results and Analysis}
\label{sec:results}

\subsection{Comparison to the State-of-the-art}
We compare our proposed framework to other approaches under different levels of supervision. 
Table~\ref{tab:breakfast} compares our work to others on Breakfast. Using FV features, our approach has 49.5\%, outperforming by 1.4\% the unsupervised method VTE~\cite{vidalmata2021joint}. 
ASAL~\cite{li2021action} adopts the same initialization as~\cite{kukleva2019unsupervised}, combined with a temporal order classification module to learn the feature embedding with updated pseudo-labels from Viterbi decoding iteratively. With FV, ASAL achieves the best performance of 52.5\% compared to CAD (ours) of 49.5\%. Strictly speaking, our results are not directly comparable to existing unsupervised approaches because we forced our globally discovered action prototypes to be evaluated at an activity level, and our problem setup is more challenging, as discussed in Table~\ref{tab:hungarian}.
With the more robust I3D feature, we achieve 53.1\%, which surpasses the best weakly supervised method CDFL~\cite{li2019weakly} by a margin of 2.9\% with the supervision of an ordered action list.  

We summarise in Table~\ref{tab:yti} the competing unsupervised approaches on YouTube. For video-level matching, the best performing approach is TW-FINCH~\cite{sarfraz2021temporally}, with a 48.2\% F1 score, which is exceedingly high since the task is relatively easier when segmenting per video. Our approach, at the activity level, has an F1 score of 35.1\%, slightly lower than the state-of-the-art~\cite{elhamifar2019unsupervised}, with 37.3\%. ASAL~\cite{li2021action} achieves the highest MoF score of 44.9\% and our approach outperforms closely related work CTE~\cite{kukleva2019unsupervised} (40.5\% compared to 39.0\%). 
\begin{table}[t]
\centering
\caption{Comparison of proposed method with other state-of-the-art approaches for fully, weakly and unsupervised learning on the Breakfast dataset. FV denotes Fisher vector.}
\label{tab:breakfast}
\scalebox{1}{
\begin{tabular}{l|l|c|c}
\hline
Supervision                                         & Approach  & Features & MoF (\%)   \\ \hline
\multirow{6}{*}{Full}                           
                                                    & HTK~\cite{kuehne2016end}   &   I3D      & 56.3  \\
                                                    & GRU~\cite{richard2017weakly} & I3D   & 60.6  \\
                                                    & MS-TCN++~\cite{li2020ms}     & I3D & 67.6  \\
                                                    & Local SSTDA~\cite{chen2020action} &I3D  & 70.2  \\
                                                    & SSTDA~\cite{chen2020action}       &  I3D  & 70.3  \\ \hline
\multirow{6}{*}{Weak}                               & Fine2Coarse~\cite{richard2016temporal}  & FV   & 33.3  \\
                                                    & GRU~\cite{richard2017weakly}          &  FV & 36.7  \\
                                                    & TCFPN+ISBA~\cite{ding2018weakly}    &FV & 38.4  \\
                                                    & NN-Viterbi~\cite{richard2018neuralnetwork} &  FV   & 43.0    \\
                                                    & D3TW~\cite{chang2019d3tw}        & FV& 45.7  \\
                                                    & CDFL~\cite{li2019weakly}        & FV& 50.2  \\ \hline
\multicolumn{1}{c|}{\multirow{5}{*}{Unsupervised}} & GMM~\cite{sener2018unsupervised}     &  FV   & 34.6  \\
\multicolumn{1}{c|}{}                              & CTE~\cite{kukleva2019unsupervised}   & FV& 41.8  \\
\multicolumn{1}{c|}{}                              & VTE~\cite{vidalmata2021joint}         &  FV & 48.1 \\ 
\multicolumn{1}{c|}{}                              & ASAL~\cite{li2021action} & FV & 52.5\\\hline

\multirow{2}{*}{CAD (Ours)}              &               & FV          & 49.5 \\
                           & & I3D         & \best{53.1} \\ \cline{2-3}  \hline
\end{tabular}}

\end{table}
\begin{table}[htb]
\centering
\caption{Comparisons with other unsupervised action segmentation works on the YouTube Instructions dataset. F1 score and MoF are reported.}
\label{tab:yti}
\begin{tabular}{l|c|c|c}
\hline
Approach         & Hungarian & F1 & MoF (\%)  \\\hline
LSTM+AL\cite{aakur2019perceptual}   & video & 39.7    & -    \\
TW-FINCH\cite{sarfraz2021temporally} & video & \best{48.2} & \best{56.7}\\
\hline
Frank-Wolfe\cite{bojanowski2014weakly} & activity & 24.4     & -    \\
Mallow\cite{sener2018unsupervised} &  activity   & 27.0     & 27.8  \\
CTE\cite{kukleva2019unsupervised}    &  activity   & 28.3     & 39.0  \\
VTE\cite{vidalmata2021joint}    &  activity   & 29.9     & -     \\
JointSeqFL~\cite{elhamifar2019unsupervised} & activity & \best{37.3} & - \\
ASAL~\cite{li2021action} & activity &32.1&\best{44.9}\\\hline
CAD (Ours)        &   activity  &  {35.1} & 40.5  \\\hline
CAD (Ours)         &    global   & 12.1 &  15.7    \\\hline
\end{tabular}

\end{table}

\subsection{Ablation Study}
\textbf{Various Settings.} 
We study and report the MoF accuracy under various settings for the Breakfast dataset in Table~\ref{tab:ablation} with both FV and I3D features. In a global matching setting with 50 prototypes, our approach achieves 10.29\% with FV and 19.2\% with I3D, and we provide this as a baseline.
We then perform the activity-level matching and compare with each complex activity having five actions versus the maximum number of actions per activity based on the ground truth. Using five actions has a 5.1\% higher MoF than the maximum (34.0\% vs 28.9\%).  
This is not surprising as a smaller number of constituent actions tends to under-segment the video; this increases the MoF value by favouring the most frequent actions, with the extreme case assigning the same dummy label for all frames. 
Adding decoding on top, we see a boost of approximately 10\%; this indicates that decoding is very helpful as it incorporates the temporal reasoning between actions. We further find that a Gaussian smoothing before decoding is also helpful and can add $10-15\%$ depending on the size of the kernel. A kernel with $\sigma\!=\!5$ leads to our best result of 53.1\% MoF.

\textbf{Loss Terms.}
Table~\ref{tab:losscomp} is a study of the different loss terms defined in Eq.~\eqref{eq:finalloss}.  Interestingly, with only visual representation $V^g$ ($\alpha=0$), our model performance is very poor, achieving 13.1\% (global) and 30.8\% (activity).
Meanwhile, with increasing $\alpha$, our model achieves much better results, and we empirically find that  $L^p$ imposed on the affinity matrix $A$ can help guide the learning. The best score is achieved with $\alpha=0.5$. The effect of the smoothing term $L^s$ is also shown in Table~\ref{tab:losscomp}. The values suggest that imposing this term helps to boost the performance marginally (approximately 1\%).

\begin{table}[t]
\centering
\caption{Ablation studies on hyper-parameters $\alpha$ and $\lambda$ in our final loss formulation.}
\label{tab:losscomp}
\scalebox{1}{\begin{tabular}{l|ccccc|cc}
\hline
\multirow{2}{*}{} & \multicolumn{5}{c|}{$\alpha$}   & \multicolumn{2}{c}{$\lambda=0.15$} \\ \cline{2-8} 
                  & 0    & 0.25 & 0.5  & 0.75 & 1    & +         & -         \\ \hline
activity          & 13.1 & 19.1 & \textbf{19.2} & 19.1 & 17.1 & \textbf{19.2}      & 18.4      \\
global            & 30.8 & 50.9 & \textbf{53.1} & 51.2 & 50.5 & \textbf{53.1}      & 51.4   \\\hline  
\end{tabular}}

\end{table}
\begin{table}[ht]
\centering
\caption{Model performances with different numbers of prototypes. MoP and MoC are mean over prototypes and mean over classes, respectively. }
\label{tab:nproto}
\begin{tabular}{l|c|c|ccc}
\hline
\multirow{2}{*}{} & \multirow{2}{*}{\begin{tabular}[c]{@{}c@{}}\# of \\ prototypes\end{tabular}} & Activity & \multicolumn{3}{c}{Global} \\\cline{3-6}
 &  & MoF & MoF & MoP & MoC\\\hline
CTE~\cite{kukleva2019unsupervised} & 82 & 41.8 & - & - & -\\\hline
\multirow{5}{*}{CAD (Ours)} & 20 & 49.4 & 17.5 & 15.8 & 6.58\\
 & 30 & 52.7 & 20.2 & 12.1 & 7.56    \\
 & 40 & 52.4 & 18.1 & 9.6 & 7.98 \\
 & 50 & \textbf{53.1} & 19.2 & 8.1 & \textbf{8.44} \\
 & 60 & 49.9 & 17.0 & 6.3 & 7.88\\\hline 
\end{tabular}
\end{table}

\textbf{Prototype Number $N$.} 
Table~\ref{tab:nproto} shows results for different numbers of prototypes. Compared to CTE~\cite{kukleva2019unsupervised}, which models a total number of 82 clusters, our framework requires fewer prototypes to achieve comparable activity-level results.

\textit{Activity} 
On the activity level, when $N\!=\!30$ and $N\!=\!40$, MoF accuracy is about the same and reaches its peak at 53.1\% with $N=50$, which is close to the ground truth number of actions (48) on Breakfast. Increasing to 60 prototypes decreases the performance (3.2\%), since any frames assigned to the extra prototypes is automatically considered wrong after the Hungarian matching. When the number of prototypes decreases to only 20, our proposed model still outperforms CTE~\cite{kukleva2019unsupervised} by a large margin (7.6\%). The overall high activity-level performances illustrate the stability of the learned prototypes, \ie, despite the changes in the global prototype number $N$, the selected $N'$ prototypes for each complex activity still demonstrate a strong capability in differentiating actions.

\textit{Global} 
A similar trend can be seen in the global performance. The MoF drops when the number diverges significantly from the ground truth (48) number of actions. When $N\!=\!60$, Hungarian matching cannot account for the unmatched actions. A smaller number of prototypes cannot sufficiently represent the range of actions, \eg, $N\!=\!20$. 

At first glance, it may seem strange that the MoF value does not change with the number of prototypes $N$, even if this deviates significantly from the ground truth. However, the cause is rooted in the fact that MoF is a poor standalone evaluation measure since it reflects only the overall accuracy over all frames without accounting for the class-wise distribution.
Once the dominant action classes are represented in the segment clusters, under-representing the tail classes in frame numbers will have little impact on the MoF value.
Therefore, we advocate using a prototype- or action-wise accuracy to better illustrate the influence of changing prototype numbers. To that end, we report the Mean over Prototypes (MoP) and Mean over Classes (MoC) in Table~\ref{tab:nproto}, averaging accuracy over prototypes and ground truth action classes, respectively. As the number of prototypes decreases, an increase in MoP is expected as fewer prototypes under-segments the video sequence. Under-segmentation results in an MoF increase (18.1\%$\rightarrow$20.2\%) as the matching of dominant action classes boosts the overall accuracy. In contrast, the MoC, which is normalized by the ground-truth number of total actions, \ie, 48 on Breakfast, better reflects the performance changes with respect to different numbers of prototypes. The highest MoC value of $8.44\%$, corresponding to  $N\!=\!50$, suggests that modelling 50 prototypes produces the best performance; adding or removing prototypes decreases the MoC.

\subsection{Unknown Complex Activity Labels}\label{subsec:unknown}
In CTE~\cite{kukleva2019unsupervised} and ASAL~\cite{li2021action}, they extend their approach into a setting that considers all complex activities and performs Hungarian matching on a `global' level. We argue that it is still not equivalent to our global setting, as discussed in Section~\ref{sec:hun}. Under their setting, where no complex activity label is known, they first run a bag of words clustering on the videos to partition them into multiple pseudo-activities. Then, they perform their action clustering within each pseudo-activity individually. In other words, they apply their activity-level action segmentation within pseudo-activity classes. Their approach still cannot accommodate possible shared actions across activities. 

To align with CTE~\cite{kukleva2019unsupervised} and ASAL~\cite{li2021action}, we adopt the identical video clustering results as pseudo-labels to supervise our classification and then do Hungarian matching. The results are reported for Breakfast in Table~\ref{tab:unsup}. The pseudo-labels adopted have a Mean over Videos (MoV) accuracy of 32.8\% through Hungarian matching between video clustering results and ground truth complex activity labels. For each pseudo-class, we also assume $N\!=\!5$ prototypes to match the five constituent actions used in~\cite{kukleva2019unsupervised}. It can be seen from the table that our approach achieves a comparable MoF of 17.7\% with our na\"{i}ve labeling and is further boosted to 23.4\% after applying the decoding. Compared to 18.5\% from CTE~\cite{kukleva2019unsupervised}, our model achieves higher performance at 23.4\%. Similarly, our approach outperforms ASAL~\cite{li2021action} by a large margin of 3.2\%. This increase is likely because the inaccurate pseudo labels cause more action sharing of video instances across pseudo activity classes, and our framework is better at discovering them.

\begin{table}[t]
\centering
\caption{Fully unsupervised action segmentation on Breakfast. `MoV' denotes mean over videos, which is the accuracy of matching between video clusters and ground truth complex activity classes. `N' indicates how many actions are considered for each activity.}
\label{tab:unsup}
\begin{tabular}{l|c|c|c|c}
\hline
Approach     & N & MoV (\%) & Decoding & MoF (\%)   \\ \hline
CTE + BoW~\cite{kukleva2019unsupervised}  & 5 & 32.8 & \cmark         & 18.5  \\ 
ASAL~\cite{li2021action} & 5 & - & \cmark & 20.2 \\ \hline
CAD (Ours) + BoW & 5 & 32.8 & \xmark         & 17.7  \\ 
CAD (Ours) + BoW & 5 & 32.8 & \cmark         & \best{23.4}  \\ \hline
\end{tabular}

\end{table}
\begin{table*}[]
\centering
\caption{KL divergence of action frame distribution between predicted and ground truth on Breakfast.  $K$ is the ground truth composing actions per activity. Compared with CTE~\cite{kukleva2019unsupervised}, our action discovery framework demonstrates a better estimation.}
\label{tab:kl}
\scalebox{1}{
\begin{tabular}{l|c|c|c|c|c|c|c|c|c|c|c}
\hline
Activity & pancake & cereal & tea  & milk & juice & sandwich & scrambled egg & friedegg & salat & coffee & Average \\\hline
K (\# of actions) & 13 & 5 & 7 & 5& 8 & 9 &  11 & 9 & 8 & 7 & - \\\hline
CTE\cite{kukleva2019unsupervised} $\rightarrow$ GT & 0.62     & 0.15   & 0.34 & 0.17 & \textbf{0.40}  & 0.37     & 0.50      & 0.53     & 0.46  & 0.23   & 0.38    \\\hline
CAD (Ours) $\rightarrow$ GT & \textbf{0.32} & \textbf{0.11} & \textbf{0.27} & \textbf{0.16} & 0.42  & \textbf{0.12} & \textbf{0.47} & \textbf{0.42} & \textbf{0.27} & \textbf{0.15} & \textbf{0.27}   \\\hline 
\end{tabular}}

\end{table*}
\begin{table*}[htb]
\centering
\caption{Top seven activated prototypes on three complex activities. These prototypes are sorted in descending order of frequency. Bold and italic \correct{items} denote the right correspondence between discovered prototypes and actual actions in the ground truth.}
\label{tab:global}
\scalebox{1}{
\begin{tabular}{l|l}
\hline
Activity & \multicolumn{1}{c}{Top Activated Prototypes}\\ \hline
pancake  & \correct{fry\_pancake} - \correct{stir\_dough} - fry\_egg - \correct{butter\_pan} - \correct{pour\_milk} - \correct{pour\_dough2pan} - \correct{take\_plate}\\ \hline
friedegg & fry\_pancake - \correct{butter\_pan} - \correct{fry\_egg} - stir\_dough - \correct{take plate} - pour\_milk - pour\_juice      \\ \hline
(fruit) salat    & \correct{cut\_fruit} - stir\_dough - \correct{peel\_fruit} - butter\_pan - crack\_egg - \correct{put\_fruit2bowl} - pour\_juice  \\ \hline
\end{tabular}}

\end{table*}
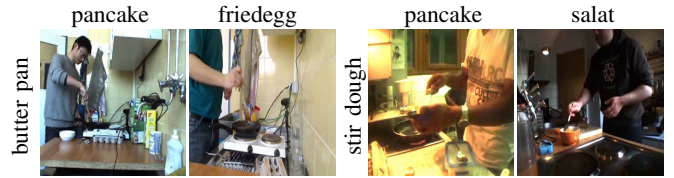
\begin{figure}[tb]
    \centering
    \begin{overpic}[width=\linewidth]{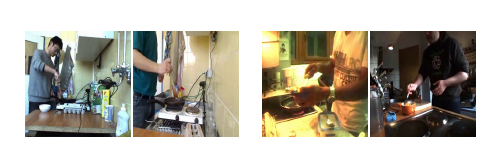}
    \put(1,5){\rotatebox{90}{\small butter pan}}
    \put(10,25){\small pancake}
    \put(32,25){\small friedegg}
    \put(51,5){\rotatebox{90}{\small stir dough}}
    \put(60,25){\small pancake}
    \put(85,25){\small salat}
    \end{overpic}
    \caption{For given prototype queries `butter pan' and `stir dough', we retrieve most activated frames from complex activities ``pancake'', ``friedegg'' and ``salat''. The retrieved images demonstrate a motion pattern similar to the provided query action.}
    \label{fig:retr}
\end{figure}
\newcommand{\rott}[1]{\rotatebox{45}{\scalebox{0.8}{\small #1}}}
\newcommand{\rotl}[1]{\rotatebox{0}{\scalebox{0.8}{\small #1}}}
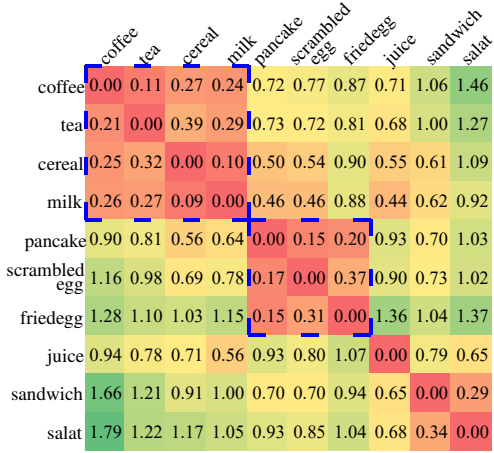
\begin{figure}[t]
    \centering
    \begin{overpic}{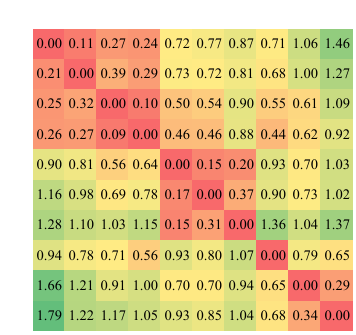}
    \put(12,86){\rott{coffee}}
    \put(21,86){\rott{tea}}
    \put(30,86){\rott{cereal}}
    \put(40,86){\rott{milk}}
    \put(46,86){\rott{pancake}}
    \put(54,86){\rott{scrambled}}
    \put(59,86){\rott{egg}}
    \put(66,86){\rott{friedegg}}
    \put(75,86){\rott{juice}}
    \put(84,86){\rott{sandwich}}
    \put(92,86){\rott{salat}}
    \put(-1,80){\rotl{coffee}}
    \put(4,71.5){\rotl{tea}}
    \put(-1,63){\rotl{cereal}}
    \put(1,54.5){\rotl{milk}} 
    \put(-8,39.5){\rotl{scrambled}}
    \put(3,36.5){\rotl{egg}}
    \put(-4,46){\rotl{pancake}}
    \put(-5,29){\rotl{friedegg}}
    \put(1,20.5){\rotl{juice}}
    \put(-7,12){\rotl{sandwich}}
    \put(1,3.5){\rotl{salat}}   
        
    \put(9.5,52){\color{blue}\linethickness{0.5mm}
        \dashline{4.5}(0,0)(0,34)(36,34)(36,0)(0,0)}
    \put(45.5,26.5){\color{blue}\linethickness{0.5mm}
        \dashline{4.5}(0,0)(0,25.5)(27,25.5)(27,0)(0,0)}
        
    \end{overpic}
    \caption{KL divergence of prototype distributions for complex activity pairs over the entire set of global prototypes on Breakfast. Lower value (red color) indicates more similarity between the two prototype distributions. Best viewed in color.}
    \label{fig:sharing}

\end{figure}

\subsection{Learned Prototypes}
Ideally, the global set of learned prototypes should have 1) adequate discriminability in segmenting actions and 2) competence in capturing shared actions across complex activities. We further investigate the former point from the activity level and the latter at the global level.

\textbf{Discriminability.}
\label{subsec:dist}
Other than the MoF, which indicates how accurate the frame-wise predictions are, the distribution of actions can also be a good indicator of the model's discriminativeness. We estimate in Table~\ref{tab:kl} on an activity level the discrepancy between the model output's frame-wise empirical distributions of actions versus the ground truth. To measure the discrepancy, we apply the KL divergence to the two distributions: 
\begin{equation}
\label{eq:kl}
    D(p||q) = \sum_{k\in K} p(k)log \frac{p(k)}{q(x)},
\end{equation} 
where $p(k)$ and $q(k)$ are, respectively, the model output and ground truth frame distributions for the same complex activity with $K$ composing actions. The frame distribution over actions is defined as:
\begin{equation}
    p(k) = \frac{\sum_{t=1}^{T} \mathds{1}(l_t=k)}{T}
\end{equation}
where $\mathds{1}(\cdot)$ is the indicator function.
Table~\ref{tab:kl} compares our divergence scores to CTE~\cite{kukleva2019unsupervised} and shows that our approach achieves lower divergences for all activities except ``juice''. Even for this exception, our divergence is only marginally higher (0.42 vs 0.40). For the most complicated activity, ``pancake'', with a total number of 13 composing actions, our approach (0.32) surpasses CTE (0.62) by the most significant margin. The lower divergences from the table demonstrate that our approach better estimates the actions' frame-wise distributions than CTE~\cite{kukleva2019unsupervised}. We posit that the advantage comes from the affinity summation over time in our prototype representation $V^p$ (see Eq.~\eqref{eq:prototype_aggregation}), which can take into consideration both the action occurrence and the frequency.  

\textbf{Shared Actions.} 
On a global level, we first try to interpret the sharedness in the global set of prototypes by summarizing three complex activities and their top seven activated composing prototypes (actions) in Table~\ref{tab:global}. It can be seen that for each complex activity, our approach can indeed discover constituent actions. For example, on ``pancake'', 5 out of 7 prototypes are correctly associated with ground truth actions. If we compare the prototypes of ``pancake'' with ``friedegg'', it is not hard to find that similar sets of prototypes are being activated; this is plausible because both of them are similar activities in the sense that they all involve cooking with a pan. Meanwhile, the set of constituent prototypes for ``(fruit) salat'' is quite different from ``pancake'' and ``friedegg'', since `cut\_fruit', `peel\_fruit' and `put\_fruit\_to\_bowl' are three ``salat'' exclusive actions, and they have been correctly identified. There are also mismatches between the prototypes. For instance, `stir\_dough' appears in ``salat'', which is unlikely. However, we note that discovered actions tend to focus on motion dynamics,  and `stir\_fruit' and `stir\_dough' follow a similar movement pattern, as demonstrated in Fig.~\ref{fig:retr}. Such ambiguity can be alleviated to some extent when we confine the performed matching to the activity level.

We can also probe the extent of sharing in the prototypes with the KL divergence. Consider redefining $p$, $q$ from prediction against ground truth in Eq.~\eqref{eq:kl} to instead represent two complex activities respectively, with $K\!=\!50$.  We plot the KL divergences of prototype distributions over the entire global set of (50) prototypes between activity pairs. Fig.~\ref{fig:sharing} shows how complex activities naturally group together based on their low divergence values (dashed blue rectangles). Such groupings indicate that our classification model indeed learns semantics and shared actions in these complex activities.
For example, ``coffee'', ``tea'', ``cereal'', and ``milk'' are semantically similar in terms of action composition as they all first pour ingredients into a food container and then brew with water. Similarly, additional distinct groupings occur with ``scrambled egg'', ``friedegg'' and ``pancake'', which share steps such as taking a pan, oiling the pan and cooking the food. The formation of these distinct groupings verifies that our learned prototypes are more shared between similar activities and less for distinctive pairs. The only outlier in the ten complex activities is ``juice''. This is likely because none of the other activities involves the unique `take\_squeezer', `cut\_orange' and `squeeze\_orange' actions. 

\begin{table}[tb]
\centering

\caption{Action recognition accuracy with different weightings on our dual video representations on Breakfast Action dataset. `Avg' denotes the average over SP1-4. SP5 is the protocol from~\cite{hussein2019timeception} which used 1357 videos for training and 335 for testing.}
\label{tab:dual}
\scalebox{1}{
\begin{tabular}{l|c|c|c}
\hline
\multirow{2}{*}{} & \multicolumn{3}{c}{Activity Recognition Accuracy (\%)}  \\ \cline{2-4}
                  & $w_p=1,w_g=0$     & $w_p=0,w_g=1$    & $w_p=w_g=0.5$                                      \\ \hline
SP1            & 80.16       & 77.38      & \best{81.35}                                       \\ \hline
SP2            & 68.74       & 70.51      & \best{70.73}                                       \\ \hline
SP3            & \best{77.83}       & 71.36      & 76.21                                       \\ \hline
SP4            & \best{75.52}       & 73.09      & \best{75.52}                                      \\ \hline
Avg(1-4)              & 75.56       & 73.08      & \best{75.95}                                     \\ \hline
SP5~\cite{hussein2019timeception}          & 78.81       & 79.10      & \best{80.51}\\ \hline
\end{tabular}}

\end{table}
\subsection{Complex Activity Recognition}

Given that our proposed CAD framework is a classification framework that uses complex activity labels for supervision, our framework can, as a byproduct, also be applied to recognize a complex activity. From the dual video representations $V^g$ and $V^p$ (see Sec.~\ref{subsec:dual}), we can predict the complex activity label $\hat{c}$ for a video by taking the MAP estimate, \ie:
\begin{equation}\label{eq:complex}
\hat{c}=\underset{c}{\mathrm{argmax}} (w_{p} \cdot y_c^p+w_{g} \cdot y_c^g) 
\end{equation}
where $w_p$ and $w_g$ are two weighting factors for the dual video representations.  

\textbf{Dual Video Representation.}
We report the performance with different video representations on the Breakfast Actions dataset in Table~\ref{tab:dual}. SP1-4 are four conventional splits from the dataset, and SP5 is the protocol used in~\cite{hussein2019timeception}, which used 1357 videos for training and 335 for testing. As we can see, on SP1, the visual representation $V^g$ ($w_g=1, w_p=0$) achieves a slightly higher performance of 80.16\% than the prototype representation $V^p$ ($w_g=0,w_p=1$) at 77.38\%. Similar trends can also be observed for other splits. Furthermore, with $w_g=w_p=0.5$, we achieve the best performance among all cases (81.35\%). However, merging the two achieves a modest gain of 1.19\%, highlighting that even though $V^g$ and $V^p$ are both derived from affinity matrix $A$, there are still some complementary aspects.

\textbf{Comparison with Previous Works.}
Regarding activity recognition, we cross-validate on the conventional splits of Breakfast and also use the non-standard single split of~\cite{hussein2019timeception, hussein2020pic} for a fair comparison with their work. The results are reported in Table~\ref{tab:actrec}. Timeception~\cite{hussein2019timeception} is specifically designed to reason temporal patterns for recognizing activities. With their own protocol and I3D features, they achieve 69.3\% accuracy and 71.25\% with stronger 3D Resnet50 features, respectively. Compared to them, either of our dual representations on its own already achieves state-of-the-art results, and combining them achieves an accuracy of 80.51\%, boosting the performance by a large margin of 11.21\%. We also achieve 75.95\% over four conventional splits, around 4.7\% higher than Timeception~\cite{hussein2019timeception} with 3D Resnet-50 features. We do not explicitly reason the temporal relations between actions in the proposed classification model, while our interpretation of such a performance boost is that compared to the temporal patterns, finding the right set of discriminative feature basis (composing actions) is more effective on the Breakfast dataset. We also include the unpublished work PIC~\cite{hussein2020pic} for comparison; we find that our approach is comparable to theirs with fine-tuned I3D features (80.51\% vs 80.64\%). 

\begin{table}[t]
\centering
\caption{Action recognition performance on the Breakfast Actions dataset. * denotes work that uses finetuned I3D features. SP1-4 denotes the cross-validation. }
\label{tab:actrec}
\begin{tabular}{l|c}
\hline
Approach                  & Acc (\%) \\ \hline
I3D                       & 64.31   \\
I3D + Timception \cite{hussein2019timeception}           & 69.30   \\
3D Resnet50               & 66.73   \\
3D Resnet50 + Timeception \cite{hussein2019timeception} & 71.25   \\ \hline
finetuned I3D \cite{hussein2020pic} * & 80.64 \\
finetuned I3D + PIC \cite{hussein2020pic} * & 89.84\\\hline
CAD (Ours) + I3D ($w_p=1, w_g=0$) & 78.81\\
CAD (Ours) + I3D ($w_p=0, w_g=1$) & 79.10\\
CAD (Ours) + I3D ($w_p=w_g=0.5$)  & \best{80.51}\\
CAD (Ours) + I3D + SP1-4 ($w_p=w_g=0.5$)  & 75.95   \\
\hline
\end{tabular}

\end{table}

\subsection{Limitations} 
The formulation of the learning action prototypes with frames based on their similarities is a permutation invariant design, and such method is agnostic of the sequential temporal information within each prototype group. Therefore, the model cannot disambiguate action pairs in reversing time order when applied to segment actions at a finer-grained level, \eg, `opening a bottle' vs. `closing a bottle'. Besides, estimating the number of latent prototypes is challenging when one has no prior knowledge of that video domain. 

Last but not least, we consider our framework, as a byproduct, to be appropriate for recognizing untrimmed activity videos rather than the conventional action recognition task of classifying trimmed clips without step-wise actions from datasets such as Kinetics~\cite{kay2017kinetics} and Something-something~\cite{goyal2017something}. In this case, it is of more importance to learn to disambiguate between action classes rather than discovering what are partially shared across them.

\section{Conclusion}
\label{sec:conclusion}

In this work, we present a novel Constituent Action Discovery (CAD) framework that finds a global set of prototypes for actions and only requires the high-level activity labels as supervision. We are also the first to provide a clear division of Hungarian matching protocols in temporal action segmentation without any action labels and show that our high-level weak supervision extends existing matching levels to a global one. Our proposed CAD framework exploits the inherent relationship between fine-grained actions and high-level activities to design a classification network. The prototypes are a set of trainable model parameters that are learned to best represent the video sequence as a whole. The discovered prototypes demonstrate state-of-the-art performance compared to unsupervised action segmentation approaches through our extensive experiments. In addition, CAD is also proven to help boost activity recognition tasks.

\bibliographystyle{IEEEtran}
\bibliography{IEEEabrv,jrnl_bib}

\end{document}